\documentclass[12pt,journal,compsoc]{IEEEtran}
  \usepackage[nocompress]{cite}

  \usepackage[dvips]{graphicx}

\usepackage[cmex10]{amsmath}

\usepackage{fixltx2e}

\begin{document}
% \title{Linear High Dynamic Range Imaging \\ in a Quasimonochromatic Light \\ Using Bayer-covered Photo Sensors }
% \title{Linear Wide Dynamic Range Imaging with Bayer-covered Assorted Pixels in a Quasimonochromatic Light}
% \title{Using Assorted Pixels of Bayer-covered Photosensors for Wide Dynamic Range Imaging}
\title{Using Spatially Varying Pixels Exposures and Bayer-covered Photosensors for High Dynamic Range Imaging}

\author{Mikhail V. Konnik
\IEEEcompsocitemizethanks{\IEEEcompsocthanksitem Mikhail V. Konnik is with Department of Solid State Physics, Moscow Engineering Physics Institute, Kashirskoe shosse 31, Moscow 115409, Russia.\protect\\
E-mail: konnik@pico.mephi.ru}
\thanks{Manuscript received .....}}

% The paper headers
\markboth{IEEE Transactions on Pattern Analysis and Machine Intelligence}%
{Shell \MakeLowercase{\textit{et al.}}: high dynamic range imaging, assorted pixels}

\IEEEcompsoctitleabstractindextext{%
\begin{abstract}
The method of a linear high dynamic range imaging using solid-state photosensors with Bayer colour filters array is provided in this paper. Using information from neighbour pixels, it is possible to reconstruct linear images with wide dynamic range from the oversaturated images. Bayer colour filters array is considered as an array of neutral filters in a quasimonochromatic light. If the camera's response function to the desirable light source is known then one can calculate correction coefficients to reconstruct oversaturated images. Reconstructed images are linearized in order to provide a linear high dynamic range images for optical-digital imaging systems.

The calibration procedure for obtaining the camera's response function to the desired light source is described. Experimental results of the reconstruction of the images from the oversaturated images are presented for red, green, and blue quasimonochromatic light sources. Quantitative analysis of the accuracy of the reconstructed images is provided.
\end{abstract}

}

\maketitle
\IEEEdisplaynotcompsoctitleabstractindextext
% \IEEEpeerreviewmaketitle

\section{Introduction}
\IEEEPARstart{G}{} rowing interest to the high dynamic range (HDR) imaging is observed recently. Such synthesized images from a set of the pictures with different exposure times are characterised by wide dynamic range. Numerous reconstruction and tone mapping methods~\cite{grossbergnayarResponseFromImages,devebecRecoverylinearity,manderspicardundigital} were proposed for digital HDR photography. 

But HDR imagery is suitable not only for photography. High dynamic range images are required for applications such as automotive systems~\cite{cmossensorautomotive120db}, vision systems~\cite{mirrorsHDR,nayaradaptiveHDR}, and hybrid optical-digital imaging systems based on ``wavefront coding'' paradigm~\cite{catheydowskinewparadigm,grachtdowskinewparadigm,konnikInputsenecorrelator}. 

The aim of this article is to show that using conventional solid-state photosensors with Bayer~\cite{bayer} colour filters array it is possibly to recover linear high dynamic range images. Because of using a quasimonochromatic light in hybrid optical-digital imaging systems, in this paper we are dealing with LEDs for the input scene illumination.

Linear high dynamic range images can be constructed using Spatially Varying pixel Exposures (SVE) technique, proposed in~\cite{mitsunagaSVErecover,sveassorted}. This technique allows to construct high dynamic range images using information from the neighbour pixels. When a pixel is saturated in the acquired image, it is likely to have a neighbour pixel that is not. Analysing the neighbour pixel's values, it is possible to construct a high dynamic range image. Such image is non-linear, hence linearization of the constructed SVE image is necessary. Linearization of a constructed SVE image is performed using correction coefficients that are obtained at the preliminary stage of calibration. 

The calibration procedure for obtaining of the camera's response funstion to the lightsource is provided. Experimental results of the oversaturated images reconstruction using the SVE technique are presented.

\section{Linear high dynamic range imaging}
First of all, the camera's response function to the desirable lightsource should be obtained. A flat-field scene in a quasimonochromatic light is taken with different exposures times for obtaining of camera's response function. During such calibration process, the camera's response to the lightsource as well as correction coefficients for SVE-images are calculated. 

If the correction coefficients are evaluated then oversaturated images, which are captured in the same light, can be reconstructed. Oversaturated images are analysed and saturated pixels are replaced using information from the neighbour pixels. Such constructed SVE-images are characterised by $\gamma$-like non-linearity. Then constructed SVE-images are linearized using correction coefficients, which were obtained at the stage of calibration. Reconstructed linear image is characterised by wide dynamic range and a linear response to the registered light.

\subsection{Preliminary calibration procedure}\label{sec:calibration}
The procedure of calibration is required for obtaining of the correction coefficients for the SVE images to be linearized. For such purpose, a light from the lightsource is passed through opal glass for elimination of the flat-field non-uniformity. Pictures of the flat-field are captured with different exposure times. In this work were used red, green, and blue LEDs driven with DC current. 

Main pixels are chosen accordingly to the dominant lightwave of a lightsource; then first and second extra pixels are selected. Each captured picture of the flat-field is decomposed to the three images according to Bayer's RGGB colour matrix. Denote by $I_{main}(i,j)$ the main pixels array, $I_{1}(i,j)$ the first extra pixels array, $I_{2}(i,j)$ the second extra pixels array. It is needed to construct an HDR image for each picture.  For the HDR image to be constructed properly it is necessary to determine the end of linearity (EOL) value for the camera's sensor. Used in this work camera is characterized by EOL value 3400 DN for 12-bit sensor. 

The algorithm for construction of the high dynamic range image is:
\begin{enumerate}
\item[1.] \textbf{If} $I_{main}(i,j)\geq EOL$, \textbf{Then} \\
$I_{main}(i,j) = EOL + I_{1}(i,j)$.
	\begin{enumerate}
	\item[1.1.] \textbf{If} $I_{1}(i,j)\geq EOL$, \textbf{Then} \\$I_{main}(i,j) = 2\cdot EOL + I_{2}(i,j)$.
	\end{enumerate}
\item[2.] \textbf{If} $I_{main}(i,j) < EOL$, \textbf{Then} \\
$I_{main}(i,j) = I_{main}(i,j)$.
\end{enumerate}
As a result, non-linear HDR images can be constructed using such algorithm. Using described algorithm at the stage of calibration, the radiometric function can be obtained.

\begin{figure}[htb]
\centerline{
\includegraphics[width=1\linewidth]{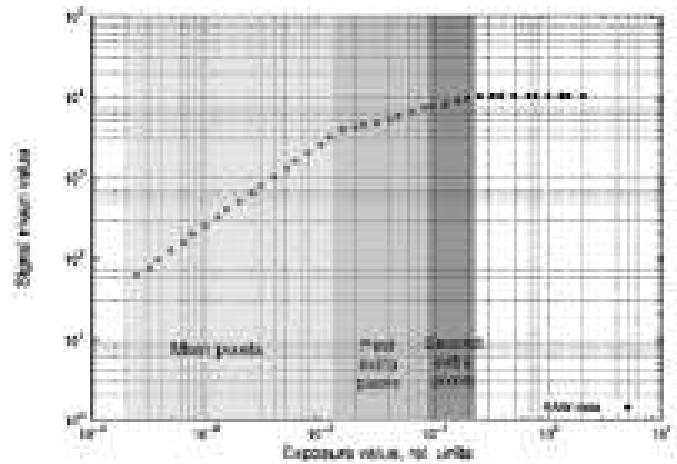}}
\caption{The effective radiometric function of the SVE imaging system.}
\label{ris:SVEresponse}
\end{figure}

\begin{figure}[htb]
\centerline{\includegraphics[width=1\linewidth]{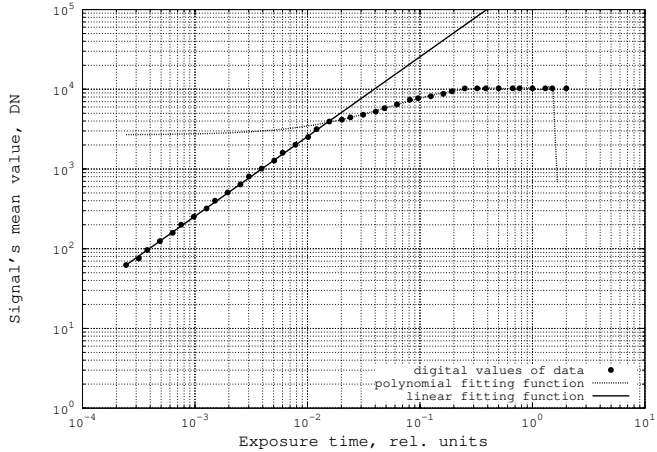}}
\caption{The approximation of the radiometric function: a line is fitted to the linear part of the data, and a high-order polynomial is fitted to the non-linear part of the data.}
\label{ris:SVEfitaccuracyMainColourREDLogscale}
\end{figure}

For the analysis and evaluation of the radiometric function, pictures of flat-field are taken for different exposures times. Then obtained images of the flat-field are constructed into HDR image using described above algorithm. An area of $256\times 256$ pixels of such image is selected, mean value of such area is calculated. Thus radiometric function, which is mean signal's value versus relative exposure time dependency, can be obtained. For example, the radiometric function for red LEDs used in this work is presented in Fig.~\ref{ris:SVEresponse}.
Further, the correction coefficients must be calculated in order to compensate non-linearity of the SVE imaging system. The linear part of the radiometric function is fitted to a line $aT + b$, where T is an exposure time (see Fig.~\ref{ris:SVEfitaccuracyMainColourREDLogscale}). The accuracy of fitting a line to the experimental data is significant: slight deviation of a line produces great errors on the reconstructed images. The Trust-Region~\cite{moretrustregionstep,byrdtrustregion} fitting algorithm was used because of speed and acceptable accuracy. Several limiting factors were specified,  for example coefficient $b$ should be the smallest possible (when there is no light, no signal should be detected).

The high-order polynomial model proposed by Mitsunaga~\cite{mitsunagaRecoverpolinoms} was used for approximation of the radiometric function's non-linear part:

\begin{equation}
	f(T) = \sum \limits_{n=0}^{N} p_n \cdot T^n
\end{equation}
As a result of the calibration, unknown coefficients $p_n$ are determined from the experimental data. Thus the correction coefficients can be calculated:

\begin{equation}\label{eq:alphasve}
	\alpha_{sve} = \frac
{aT+b}
{\sum \limits_{n=0}^{N} p_n \cdot T^n}
\end{equation}
Using Eq.\ref{eq:alphasve} for calculation of the correction coefficients $\alpha_{sve}$ it is possible to compensate SVE imaging system's non-linearity.

\subsection{Reconstruction of the SVE image}\label{sec:SVEreconstructionprocess}
Main principle of the SVE technique according to~\cite{mitsunagaSVErecover,sveassorted} is to simultaneously sample the spatial and exposure dimensions of  the image irradiance. An array  of neutral filters on a photosensor can be used to detect different levels of the input scene's signal. On the captured image, the brighter pixels have greater exposure to image irradiance and the darker ones have lower exposure. Thus the registered image contains information about the input scene signal as well as the exposure time information.

\begin{figure}[htb]
\centerline{
\includegraphics[width=0.9\linewidth]{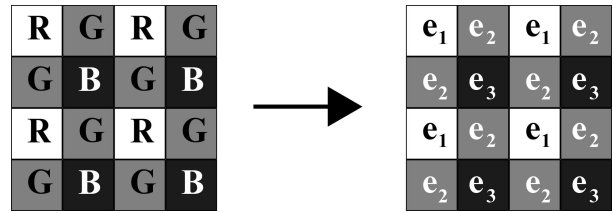}}
\caption{Bayer mosaic in a quasimonochromatic light can be considered as an array of neutral filters.}
\label{ris:SVEbayerexample}
\end{figure}

In most conventional digital cameras, Bayer's colour filters array is applied. In a quasimonochromatic light Bayer matrix can be considered~\cite{konnikGraphicon2008} as an array of neutral filters (see Fig.~\ref{ris:SVEbayerexample}). Hence it is possible to apply SVE technique for increasing of the image's dynamic range.

Denote by ``main pixels'' all pixels under filters of the colour same as illumination's colour.  If main pixels are saturated, it is likely to have neighbour pixels that are not. Denote by ``first extra pixels'' and ``second extra pixels'' neighbour pixels under different colour filters. For example, if the main pixels are red then first extra pixels are green and second extra pixels are blue.

The algorithm for SVE image construction is the same as for calibration (see Subsection~\ref{sec:calibration}). To compensate $\gamma$-like non-linearity of such SVE imaging systems, the linearization of the SVE-images is required.

For linearization of the SVE image, correction coefficients must be used. Such coefficients were calculated at the preliminary stage of calibration, as it mentioned in Subsection~\ref{sec:calibration}. The problem is that there are only a few correction coefficients that conforms to non-linear data points. In other words, for the SVE image to be linearized there are insufficient amount of correction coefficients.

The polynomial model is used for interpolation of the correction coefficients in order to overcome such problem. High-order polynomial is fitted to the data obtained at the calibration stage. Thus an unknown correction coefficient can be calculated for almost any non-linear data value of the SVE constructed image.

It is significant to estimate the accuracy of the reconstructed images due to complexity of the reconstruction process. The quantitative results of the reconstruction and linearization of the SVE images are provided below.

\section{Experimental results}
\begin{figure}[htb]
\centerline{
\includegraphics[width=1\linewidth]{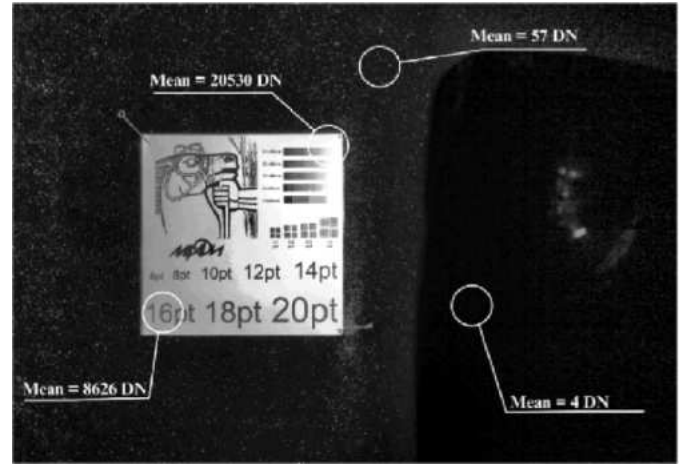}}
\caption{The test scene with high dynamic range registered by the photo sensor in the red LEDs illumination.}
\label{ris:HDRscene}
\end{figure}

The high dynamic range scene was created for the optical experiments. The photo of the test scene is presented in Fig~\ref{ris:HDRscene} (image is scaled down to 8-bit and contrasted for publishing). Scene's background is a light-absorption fabric, and the test image is illuminated by LED lamp. 

The properties of the lightsources used in this work as well as transmittance coefficients are described in Table~\ref{tab:lightsources}. It should be noted that transmittance coefficients for Bayer mosaic are obtained for used in this work commercial digital camera Canon EOS 400D.

\begin{table}[ht]
	\begin{tabular}[ht]{|l|c|c|c|}
\hline
Parameters & Main pix.& $1^{\mbox{st}}$ extra pix.& $2^{\mbox{nd}}$ extra pix.\\
\hline
%% red light %IMG_0018
\multicolumn{4}{|c|}{$\lambda_{D} = 625$ nm } \\
\hline
Pixels order &	red&	green&	blue \\
Transmittance & $e_1 = 1.00 $ &	$e_2 = 0.20$ &	$e_3 = 0.09$ \\
\hline
%% green light %IMG_0093
\multicolumn{4}{|c|}{$\lambda_{D} = 520$ nm } \\
\hline
Pixels order &	green &	blue &	red\\
Transmittance & $e_1 = 1.00 $ &	$e_2 = 0.33$ &	$e_3 = 0.15$ \\
\hline
%% blue light %IMG_0039
\multicolumn{4}{|c|}{$\lambda_{D} = 470 $ nm } \\
\hline
Pixels order &	blue &	green &	red\\
Transmittance & $e_1 = 1.00 $ &	$e_2 = 0.45$ &	$e_3 = 0.08$ \\
\hline
% red:
% 3439 - 1
% 683 - 2
% 293 - 3
% 
% green:
% 2191 -1
% 715-2
% 324 -3
% 
% blue:
% 3711-1
% 1664-2
% 305-3
	\end{tabular}
\caption{Transmittance coefficients of Bayer-covered photosensor for different LEDs used in this work.}
\label{tab:lightsources}
\end{table}

The test image consists of binary graphics, periodical elements, textual elements of different size, and gradient bars. Gradient bars are used for the estimation of the halftone stability of the reconstructed images.  The test image was captured by the digital camera with an exposure time varied from 1/4000 to 2 seconds. All captured images were processed by DCRAW~\cite{davecoffin} converter in the ``document mode'' with command \textit{dcraw -4 -D -T filename.cr2}. This command produces an unprocessed (``totally raw'') linear 12-bit TIFF images.

If an exposure time is short then a digital signal's value of the captured image is lesser than camera's ADC limit. In this case the values of the captured image are unchanged. The images captured with long exposure time are oversaturated in the main colour. Thus using information from the neighbour pixels it is possible to reconstruct such oversaturated image. The SVE reconstruction algorithm is the same as mentioned in Section~\ref{sec:calibration}.

Selecting only main pixels from the original image (``red'' pixels because of red illumination), one can obtain a new image. Let us denote such image as ``original image'' in order to calculate the NRMS error.

As above, using SVE reconstruction method and obtained calibration coefficients it is possible to reconstruct a linear high dynamic range image from the oversaturated one. The whole size of the reconstructed linear SVE-image is $1953 \times 1301$ pixels, the informative part is $300 \times 300$ pixels.

It is significant to estimate the halftone's stability of the SVE-reconstructed image. On the SVE-reconstructed image, the gradient bar with 8 halftone was analysed and 8 areas were selected in this gradient bar. Each size of the area was $15 \times 15$ pixels. To estimate the halftones stability, relations between the mean value of each area and the mean value of the ``white'' area (the brightest area) were measured. This operation was applied to every reconstructed image.

\subsection{Red LEDs illumination}
\begin{figure}[htb]
\begin{center}
	\includegraphics[width=1\linewidth]{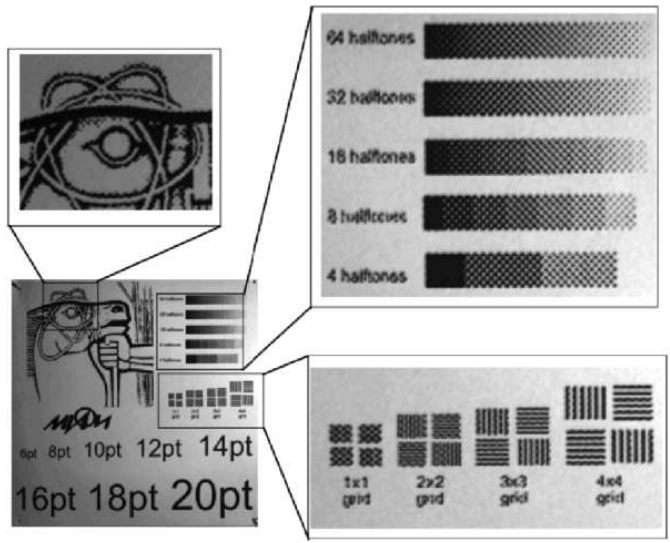}
	\caption{The reconstructed linear SVE image taken by digital camera in a quasimonochromatic red light.}
	\label{ris:FORVIEWIMG0013SVE}
\end{center}
\end{figure}

Test scene was illuminated by red LEDs with dominant wavelength $\lambda_D = 625$ nm. Main pixels for such illumination are red, first extra pixels are green, and second extra pixels are blue. As it was mentioned in Table~\ref{tab:lightsources}, transmittance  coefficient for green colour filters is 0.20 and for blue colour filters is 0.09. One of the reconstructed image is presented in Fig.~\ref{ris:FORVIEWIMG0013SVE}.

\begin{figure}[ht]
\centerline{\includegraphics[width=1\linewidth]{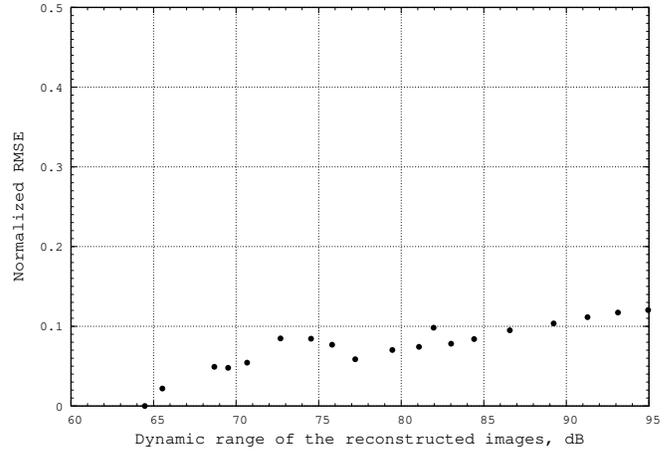}}
\caption{For red LEDs illumination: NRMS error between the original image and reconstructed images versus dynamic range.}
\label{ris:SVERMSIMG_0003experimentalhorsetoDBsignalRED}
\end{figure}

Reconstructed images using only first extra pixels are characterised by linear dynamic range of 71-84~dB and the NRMS error between the original image and reconstructed images of 5-10\% (see Fig.~\ref{ris:SVERMSIMG_0003experimentalhorsetoDBsignalRED}). Such NRMS error is considered as acceptable for practical applications in optical-digital imaging systems. For the reconstruction process there were used around 87\% of first extra pixels. 

Using first and second extra pixels it is possible to reconstruct images with dynamic range of 87-95~dB. The NRMS error between the original image and reconstructed images is around 11-15\%. There were used 96-98\% of the first extra pixels and 65-70\% of the second extra pixels to reconstruct such oversaturated images. Results of the NRMS error are summarized in Fig.\ref{ris:SVERMSIMG_0003experimentalhorsetoDBsignalRED}. 

\begin{figure}[ht]
\centerline{
\includegraphics[width=1\linewidth]{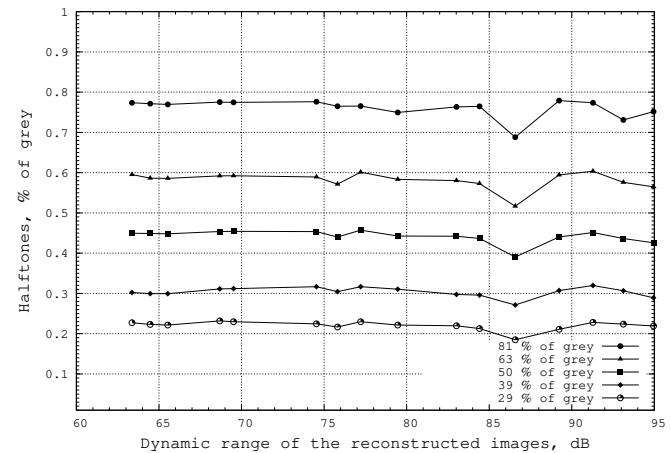}}
\caption{The halftones stability of the SVE-reconstructed images for red LEDs light.}
\label{ris:HalftoneStabilitySVEexperimentalhorseRED}
\end{figure}

The halftone stability of the reconstructed images was evaluated as well. From Fig.~\ref{ris:HalftoneStabilitySVEexperimentalhorseRED} it can be noted that images with dynamic range more than 84~dB are characterised by less stable halftone relations. Instability of the halftone relations in the range of 85 to 90~dB can be explained by transition to the second extra pixels usage. It also should be noted that halftones on the red-illuminated images are more dense, i.e., recovered image became darker than the original one. Experimental results for red LEDs illumination are summarized in Table~\ref{tab:redSVEresults}.

\begin{table}[ht]
	\begin{tabular}[ht]{|p{0.25\linewidth}|p{0.15\linewidth}|p{0.2\linewidth}| p{0.2\linewidth}|}
\hline
Parameters & Main pixels& $1^{\mbox{st}}$ extra pixels& $2^{\mbox{nd}}$ extra pixels\\
\hline
Dynamic range	& $\leq$ 70~dB	&71-84~dB	&87-95~dB\\
NRMSe		& $\leq$ 5\% 		& 5-10\% 	& 11-15\% \\
Pixels used 	& $\leq$ 1\%		& $\leq$ 87\% ($1^{\mbox{st}}$)	& $\leq$ 95\% ($1^{\mbox{st}}$) 							$\leq$ 81\% ($2^{\mbox{nd}}$) \\
\hline
	\end{tabular}
\caption{Quantitative results for SVE reconstructed images with red LEDs illumination.}
\label{tab:redSVEresults}
\end{table}

It is clear that dynamic range of the images in a quasimonochromatic red light can be increased using SVE technique up to 95~dB with acceptable (about 15\%) NRMS error from the original image and stable halftone relations on the recovered image.

\subsection{Green LEDs illumination}
The same test scene was illuminated by green LEDs with $\lambda_D = 520$ nm dominant wavelength. Main pixels for such illumination are green, first extra pixels are blue, and second extra pixels are red. According to Table~\ref{tab:lightsources}, the coefficient of transmittance for blue colour filters is 0.33 and for red colour filters is 0.15. 

It should be noted that the choice of extra pixels is ambiguous for green light illumination because of transmittance characteristics of the colour filters. When the spectrum of green lightsource is more ``cold'' then first extra pixels should be blue and second ones should be red. Otherwise when the spectrum of green lightsource is more ``warm'', it is better to assign first extra pixels as ``red'' and second ones as ``blue''. It is advisable to preliminary estimate the transmittance of green light from desired lightsource for red and blue colour filters. Wrong choice of the extra pixels can lead to halftone degradation on the reconstructed images.

Last but not the least, it is hard to obtain wide dynamic range reconstructed images in green light due to similar  coefficients of transmittance for red and blue colour filters.

\begin{figure}[htb]
\centerline{\includegraphics[width=1\linewidth]{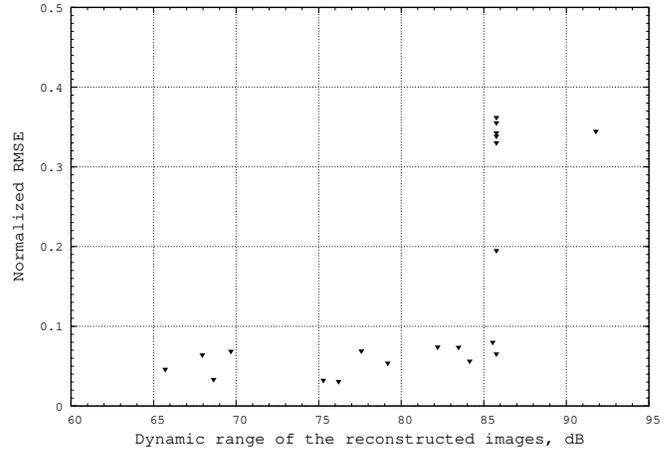}}
\caption{For green LEDs illumination: NRMS error between the original image and reconstructed images versus dynamic range.}
\label{ris:SVERMSIMG_0003experimentalhorsetoDBsignalGREEN}
\end{figure}

Reconstructed images using only first extra pixels (blue in this case) are characterised by  linear dynamic range of 75-80~dB and the NRMS error between the original image and reconstructed images of 3-7\%  (see Fig.~\ref{ris:SVERMSIMG_0003experimentalhorsetoDBsignalGREEN}). Less than 55\% of first extra pixels was used for reconstruction. 

\begin{figure}[htb]
\begin{center}
	\includegraphics[width=0.7\linewidth]{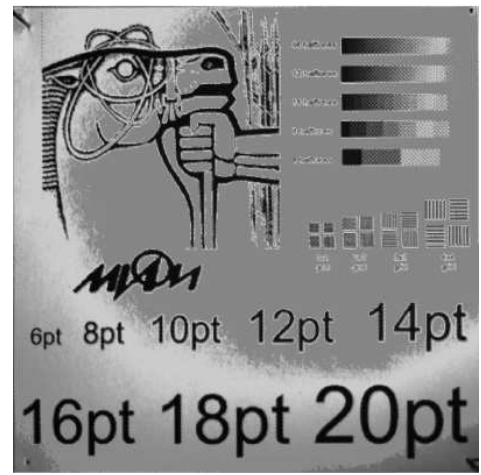}
	\caption{Reconstructed image in green LEDs illumination with wrong halftone relations.}
	\label{ris:FORVIEW-IMG_0088SVE-wronghalftones}
\end{center}
\end{figure}

\begin{figure}[htb]
\centerline{
\includegraphics[width=1\linewidth]{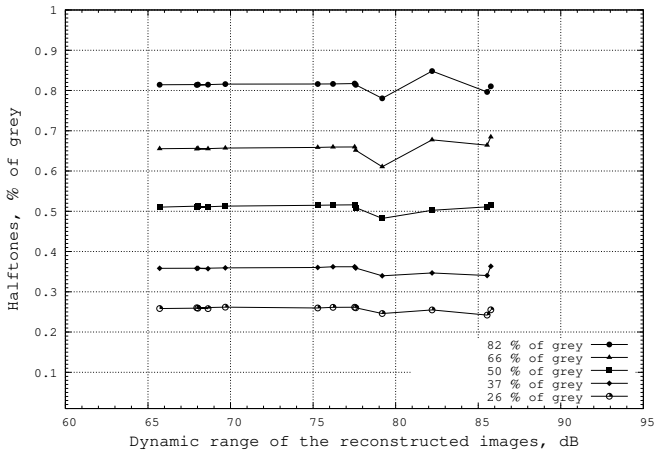}}
\caption{The halftones stability of the SVE-reconstructed images for green LEDs light.}
\label{ris:HalftoneStabilitySVEexperimentalhorseGREEN}
\end{figure}

But when second extra pixels are used there are observed significant NRMS error and halftones destabilisation (see Fig.~\ref{ris:FORVIEW-IMG_0088SVE-wronghalftones} and Fig.~\ref{ris:HalftoneStabilitySVEexperimentalhorseGREEN}). Although the dynamic range of such reconstructed images is more than 85~dB, the NRMS error is 20-35\%. Thus for the green light is needed more sophisticated algorithm in order to provide better images stability. As it mentioned above in this subsection, it is difficult to achieve a linear dynamic range of the images more than 85~dB in a green light.

\begin{table}[ht]
	\begin{tabular}[ht]{|p{0.25\linewidth}|p{0.15\linewidth}|p{0.2\linewidth}| p{0.2\linewidth}|}
\hline
Parameters & Main pixels& $1^{\mbox{st}}$ extra pixels& $2^{\mbox{nd}}$ extra pixels\\
\hline
Dynamic range	& $\leq$ 70~dB		&75-80~dB	&84-86~dB \\
NRMSe		& $\leq$ 7\% 		& 3-7\% 	& 7-38\% \\
Pixels used 	& $\leq$ 1\%		& $\leq$ 55\% ($1^{\mbox{st}}$)	& $\leq$ 91\% ($1^{\mbox{st}}$) 							$\leq$ 72\% ($2^{\mbox{nd}}$) \\
\hline
	\end{tabular}
\caption{Quantitative results for SVE reconstructed images with green LEDs illumination.}
\label{tab:greenSVEresults}
\end{table}

Obtained experimental results for green light, which are summarized in Table~\ref{tab:greenSVEresults}, allow to argue that using SVE technique it is possible to reconstruct oversaturated images to linear high dynamic range images with dynamic range up to 80~dB and NRMS error less than 7\%. However further increasing of dynamic range is required more sophisticated algorithm for image's reconstruction.

\subsection{Blue LEDs illumination} 
Test scene was illuminated by blue LEDs with dominant wavelength of $\lambda_D = 470$ nm. Main pixels for such illumination are blue, first extra pixels are green, and second extra pixels are red. As it mentioned in Table~\ref{tab:lightsources}, coefficient of transmittance for green colour filters is 0.45 and for red colour filters 0.08.

\begin{figure}[htb]
\begin{center}
	\includegraphics[width=0.7\linewidth]{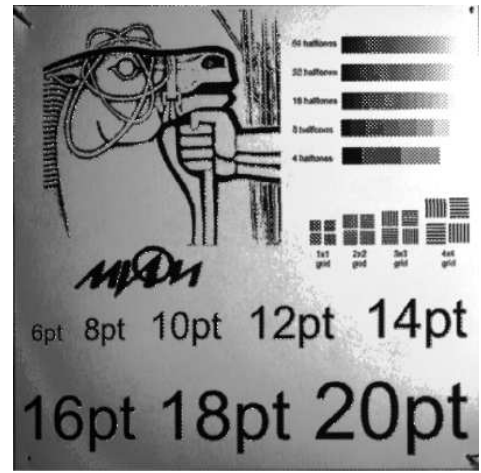}
	\caption{Reconstructed image in blue LEDs illumination.}
	\label{ris:FORVIEW-IMG_0129SVE-wronghalftones}
\end{center}
\end{figure}

Images were reconstructed using only first extra pixels (green in this case). Reconstructed images are characterised by linear dynamic range of 70-88~dB and NRMS error between the original image and reconstructed images of 9-15\%. Such NRMS error is large enough and may lead to degradation of the reconstructed image. In Fig~\ref{ris:FORVIEW-IMG_0129SVE-wronghalftones} is presented recovered image with bright spots (probably due to parasitic reflection from the laser printer's toner of the printed test image). Less than 58\% of first extra pixels were used for the reconstruction.

\begin{figure}[htb]
\centerline{\includegraphics[width=1\linewidth]{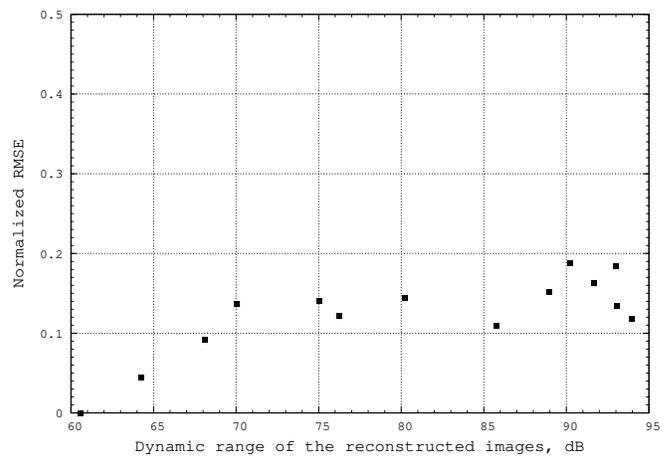}}
\caption{For blue LEDs illumination: NRMS error between the original image and reconstructed images versus dynamic range.}
\label{ris:SVERMSIMG_0003experimentalhorsetoDBsignalBLUE}
\end{figure}

\begin{figure}[htb]
\centerline{
\includegraphics[width=1\linewidth]{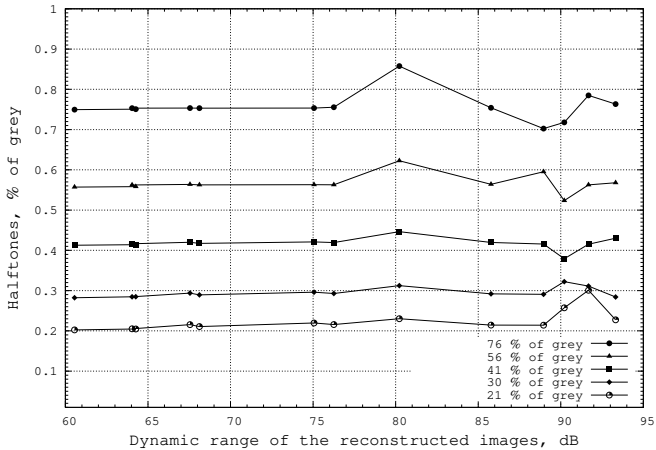}}
\caption{The halftones stability of the SVE-reconstructed images for blue LEDs light.}
\label{ris:HalftoneStabilitySVEexperimentalhorseBLUE}
\end{figure}

Using first and second extra pixels it is possible to reconstruct images with dynamic range of 90-95~dB. The NRMS error between the original image and reconstructed images is around 11-18\% (see Fig.~\ref{ris:SVERMSIMG_0003experimentalhorsetoDBsignalBLUE}). There were used 94\% of the first extra pixels and 88\% of the second extra pixels to reconstruct such oversaturated images. From Fig.~\ref{ris:HalftoneStabilitySVEexperimentalhorseBLUE} it can be noted that images with dynamic range more than 88~dB are characterised by less stable halftone relations.

Further increase of dynamic range of the images beyond 90~dB using the secondary extra pixels produces images with less stable halftone relations and bigger NRMS error from the original image. 

\begin{table}[ht]
	\begin{tabular}[ht]{|p{0.25\linewidth}|p{0.15\linewidth}|p{0.2\linewidth}| p{0.2\linewidth}|}
\hline
Parameters & Main pixels& $1^{\mbox{st}}$ extra pixels& $2^{\mbox{nd}}$ extra pixels\\
\hline
Dynamic range	& $\leq$ 70~dB		&70-88~dB	&90-95~dB \\
NRMSe		& $\leq$ 10\% 		& 9-15\% 	& 11-18\% \\
Pixels used 	& $\leq$ 1\%		& $\leq$ 58\% ($1^{\mbox{st}}$)	& $\leq$ 94\% ($1^{\mbox{st}}$) 							$\leq$ 88\% ($2^{\mbox{nd}}$) \\
\hline
	\end{tabular}
\caption{Quantitative results for SVE reconstructed images with blue LEDs illumination.}
\label{tab:blueSVEresults}
\end{table}

It can be noted that using first extra pixels one can reconstruct oversaturated images to linear high dynamic range images with dynamic range up to 88~dB and NRMS error less than 15\% (see Table~\ref{tab:blueSVEresults}). Increasing dynamic range using first and second extra pixels can produce images with less stable halftone.

\section{Conclusion} 
The reconstruction accuracy and stability of the reconstructed linear high dynamic range images using SVE technique are discussed in this paper. It is shown the possibility of the linear HDR imaging in a quasimonochromatic light using  Bayer-covered photo sensor and the SVE technique. 

For the red LEDs illumination it was obtained that SVE-reconstructed images are characterized by dynamic range up to 84~dB using first extra pixels and NRMS error between the original image and reconstructed images up to 10\%. Using both first and second extra pixels it is affordable to reconstruct a linear HDR image with dynamic range of 90-95~dB but with greater NRMS error of 11-15\%.

Results for the blue light illumination are similar to the results for red LEDs. Applying first extra pixels to restore image provides linear dynamic range of the images up to 88~dB and NRMS error up to 15\%. For both first and second extra pixels linear dynamic range increases to 90-95~dB as well as NRMS error of 11-18\%. 

Reconstructed images for the green light illumination showed lower dynamic range: only 75-80~dB for first extra pixels and NRMS error up to 7\%. Further increasing of dynamic range leads to image degradation and unacceptable NRMS error. Hence for the green light is needed more sophisticated algorithm in order to provide better images stability.

Provided quantitative analysis of the reconstructed images allows to say that the accuracy of the method is enough for practical applications. Such applications can be optical-digital imaging systems based on ``wavefront coding'' paradigm, in-vehicle systems of images understanding, and edge detection systems.

\appendices

% use section* for acknowledgement
\ifCLASSOPTIONcompsoc
  % The Computer Society usually uses the plural form
  \section*{Acknowledgments}
\else
  % regular IEEE prefers the singular form
  \section*{Acknowledgment}
\fi

This research was partly supported by the Ministry of education and science of Russian Federation (Program ``The development of the scientific potential of High School'', project RNP.2.1.2.5657).

\begin{IEEEbiography}[{\includegraphics[width=1in,height=1.25in,clip,keepaspectratio]{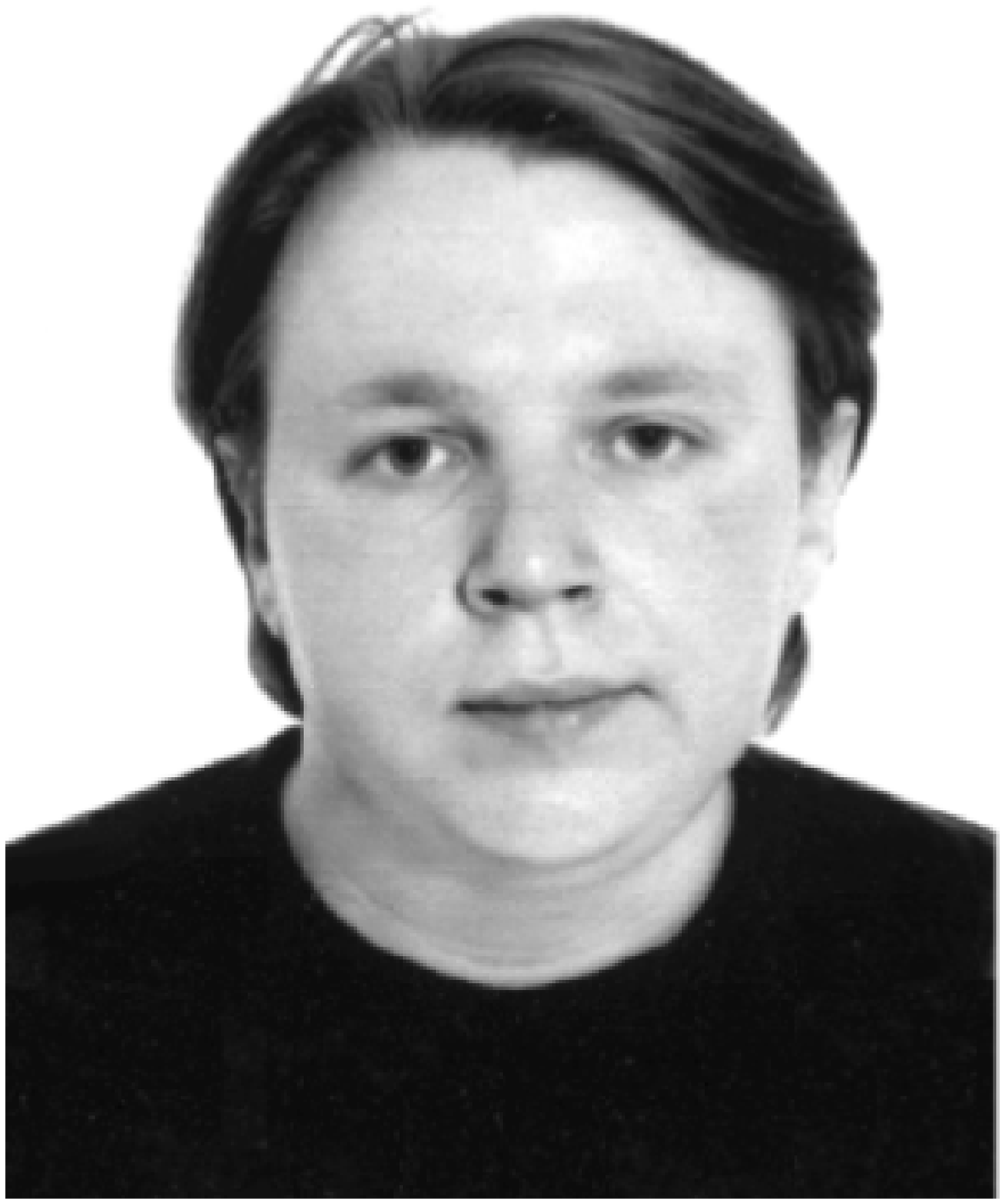}}]{Mikhail V. Konnik}
was born in Moscow, Russia, in 1982. He received the Dipl.-Ing. degree in laser physics from Moscow Engineering Physics Institute in 2006. Since 2006, he has been a Research Fellow of the
Optical Target Detection Lab, Laser Physics Department, Moscow Engineering Physics Institute. His research interests includes optical encryption, hybrid optical-digital imaging systems, image processing, and high-dynamic range imaging. Mr. Konnik is the recipient of the UMNIK Fellowship for Scientific Innovations since 2008, and received the Outstanding Paper Awards in 2005, 2007, and 2008 from the Moscow Telecommunicational Conference.
\end{IEEEbiography}

\end{document}